# Generalization in medical AI: a perspective on developing scalable models

Eran Zvuloni[1] and Leo Anthony Celi[2,3,4] and Joachim A. Behar[1,*]

[1]Faculty of Biomedical Engineering, Technion Israel Institute of Technology, Haifa, Israel,

[2]Institute for Medical Engineering and Science, Massachusetts Institute of Technology, Cambridge, MA, USA,

[3]Department of Biostatistics, Harvard TH Chan School of Public Health, Boston, MA, USA,

[4]Department of Medicine, Beth Israel Deaconess Medical Center, Boston, MA, USA.

* Corresponding author: jbehar@technion.ac.il

**Abstract:**

The scientific community is increasingly recognizing the importance of generalization in medical AI for translating research into practical clinical applications. A three-level scale is introduced to characterize out-of-distribution generalization performance of medical AI models. This scale addresses the diversity of real-world medical scenarios as well as whether target domain data and labels are available for model recalibration. It serves as a tool to help researchers characterize their development settings and determine the best approach to tackling the challenge of out-of-distribution generalization.

**Article:**

In recent years, research has witnessed the advancement of machine learning (ML) models trained on large datasets, some even encompassing millions of examples.[1,2] While these models have demonstrated impressive evaluation performance on their local hidden test sets, they often underperform when assessed on external datasets.[3–6] In light of the critical role of generalization in medical AI development, scientific journals now often require reporting on both local hidden and external datasets. Effectively, the field of medical AI has transitioned from the traditional usage of a single dataset that is split into train and test sets, to a more comprehensive framework using multiple datasets. This framework separates between "source domain" datasets used for model development and "target domain" datasets used for model evaluation.

"Generalization performance" is broad term often used to refer to the model performance on new, unseen data sourced from the same distribution as the training data. In contrast, the term "out-of-distribution generalization performance" (OOD-GP) refers to a model's performance on data from a different distribution than the training data. The decreased performance of AI models on external test sets can often be attributed to data drift. This refers to changes in the distribution of data to which the model is applied. Such drifts can stem from variations in data acquisition, such as hardware, resolution and lead placement, or biases in the training population, including patient comorbidities, ethnicity, age and sex. In addition, shortcut learning,[7] where a model learns irrelevant and coincidental relationships, can lead to satisfactory local test set performance, but will fail to transfer to real-world scenarios.

OOD-GP is essential for ensuring AI model robustness in real-world medical applications, where data distribution shifts are common. Some argue that universally OOD generalizable systems may be unattainable.[8] Given this challenge, pragmatic solutions can be categorized based on their algorithmic complexity, deployment feasibility, and effectiveness in handling various data distribution shifts. The choice of approach also depends on its stage in the model training pipeline, whether during data preprocessing, model training, or fine-tuning after deployment. One common strategy is to selectively curate training datasets, reducing bias and explicitly defining limitations by excluding certain populations.[9] Alternatively, generalization can be improved at the algorithmic level, for example, through semi-supervised re-calibration, which may eliminate the need for transfer learning when reference labels are unavailable in target domains.[10] Another approach is to move away from static external validation and instead adopt a dynamic framework that continuously updates using locally collected data.[11]

Given the trade-offs of each approach, the decline in model performance on OOD datasets remains a major challenge for widespread integration of AI tools in clinical practice. In this perspective, we introduce a three-tiered scale system that characterize the varying OOD-GP qualities and characteristics of a given medical AI model.

**Level 1: Limiting the effects of data drift or shortcut features by external validation over multiple retrospective datasets**

While developing ML models, it is critical to evaluate their *in silico* OOD-GP. The availability of multiple retrospective datasets enables researchers to develop models that are less sensitive to data drift or learn shortcut features. The simplest technique is to use multi-source domain training to train the model to learn robust cross-domain features and to evaluate the model's performance on target datasets.[12,13] Recent advancements in foundation models using self-supervised learning have also improved OOD-GP by creating better data-type representations through model pre-training on large and diverse datasets.[14,15] *In silico* OOD-GP evaluation requires curated and openly accessible datasets for specific data types. Initiatives like the National Sleep Research Resource[16] and PhysioNet[17] provide valuable datasets of this nature. However, datasets are often scattered across different repositories and may lack standardization, with labels that can vary based on differences in clinical guidelines and annotation processes. Harmonizing dataset structure, format and clinical codes is necessary for effective evaluation.

**Level 2: Unsupervised domain adaptation for model recalibration when no target domain labels are available**

Although a multi-source domain training approach may increase model robustness, Level 1 is still limiting if the goal is to capture the full range of possible distributions across any given dataset. Moreover, in practice, datasets used to develop ML models are often biased and do not accurately represent the target population. In fact, sometimes the curation process itself may introduce a bias resulting in poor representation of real-world variability. These limitations highlight the need for AI solutions that can be recalibrated for new hospital or use cases. However, even if data from a new domain are available, reference labels might be missing.

In unsupervised domain adaptation settings, labeled data are available from the source domain, while unlabeled data are used from the target domain during training. This mirrors practical medical AI deployment, where labeled datasets are used for model development, but only unlabeled data are available when the model is deployed in a new hospital setting. It is reasonable to expect continuous data collection from the hospital, but it is unlikely that labeled data will be consistently available. Similarly, in remote health monitoring systems, like smartwatches or ECG patches, recorded data are collected but not labeled. In this context, deep unsupervised domain adaptation[10] offers a promising approach to adapt a model, initially developed with source domain data, to a specific target domain that reflects the hospital's practices and policies.

**Level 3: Recurring local validation when local target domain data and labels are available**

While domain adaptation is generally a promising approach, with the potential to improve as AI techniques advance, its effectiveness diminishes when there are large distribution shifts between the source and target domains. Moreover, Levels 1 and 2 work under the widely accepted assumption[18–20] that external validation serves as the ultimate test of a model's generalizability, and that a locally developed model failing this test is not viable. Level 3 adopts a different paradigm.

External validation alone has several limitations when assessing the utility of ML systems in healthcare. It cannot fully capture data heterogeneity across geographic regions and healthcare facilities, nor does it adequately evaluate clinical usefulness and fairness, which depend on how model recommendations translate into tests or treatments within specific institutions. Additionally, data can vary over time due to both internal (e.g., EMR updates) and external (e.g., shifts in clinical practice and regulations) factors. Recognizing these challenges, Futoma et al.[8] argued that instead of cross-institutional validation, the focus should be on understanding when, how and why a system works, as ML models are context-dependent and may fail outside their original scope. Building on this idea, Youssef et al.[11] proposed recurring local validation, where models are continuously updated using new institution-specific data. This dynamic approach enhances resilience to the evolving nature of healthcare, ensuring that models remain clinically relevant while maintaining fairness and utility over time.

**Acknowledgements**

We acknowledge the financial support of the VATAT Fund to the Technion Artificial Intelligence Hub (Tech.AI) and the MISTI MIT - Israel Zuckerman STEM Fund.

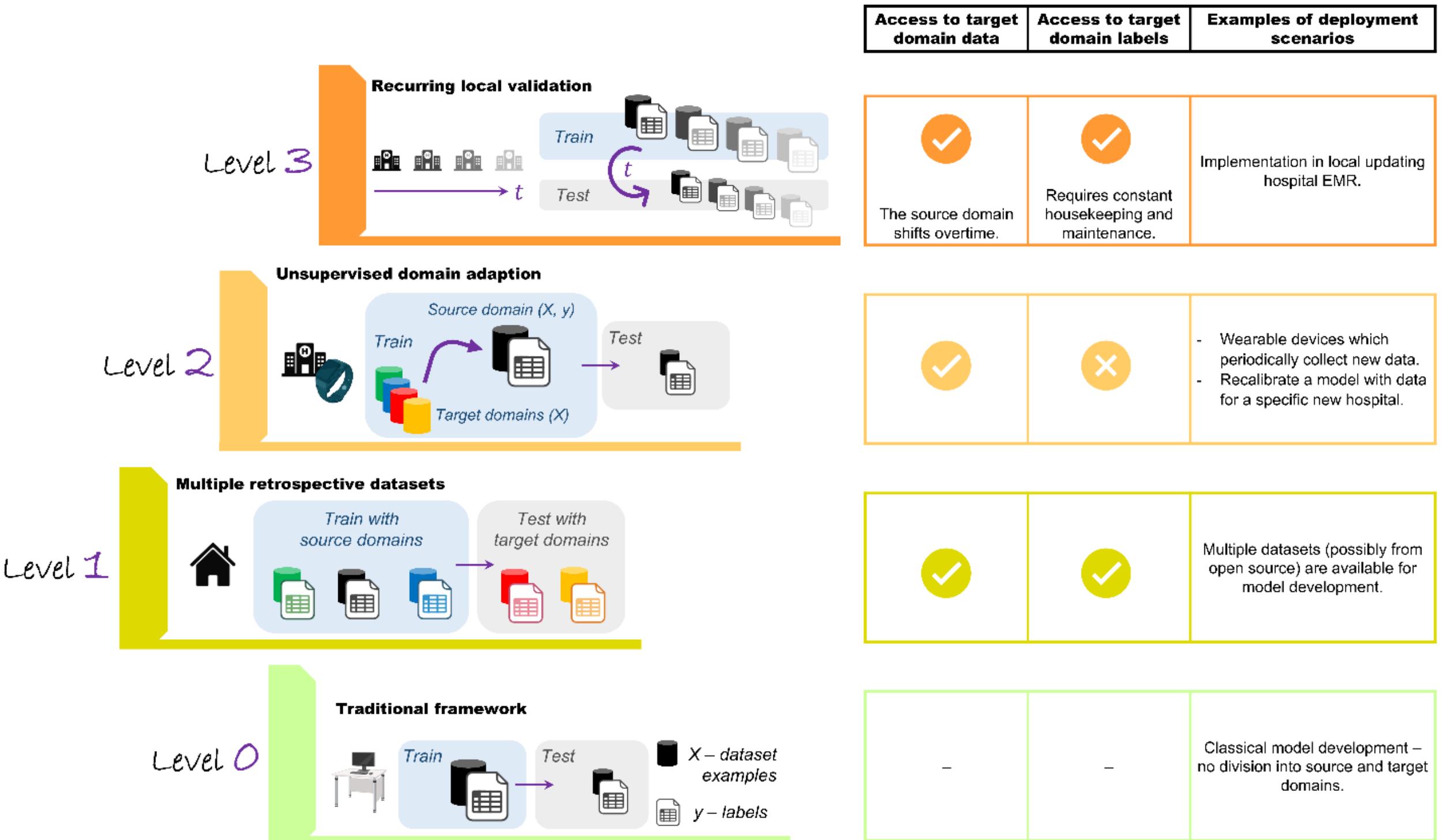

**Figure 1 | The three-level scale system reflecting the out-of-distribution generalization level of a medical AI algorithm.** From bottom to top: **Level 0 – Traditional framework.** A single dataset is split to train and test sets. The model is developed with the train set and evaluated on the test set. Both sets include dataset examples (X) and labels (y). **Level 1 – Multiple retrospective datasets.** Multiple datasets are split into and represent the source and target domains. The source datasets are used for training while the target datasets for testing. This process occurs while the model is being developed. **Level 2 – Unsupervised domain adaptation.** The labels are available only for one dataset, yet examples are available from multiple datasets. Unsupervised learning techniques are used to utilize the missing-label examples while training, to improve performance on the test set. This process is mostly suitable when data can be acquired continuously (e.g., in hospitals or via smartwatches). **Level 3 – Recurring local validation.** The process is similar to Level 0, yet the dataset changes over time and the development and evaluation process require to be repeated, adapting the model in a continuous manner. This process is suitable to take place on the site of data collection (e.g., hospital).